\documentclass{article}

\usepackage{url}
\usepackage[numbers]{natbib}
\usepackage{graphicx}
\usepackage{authblk}
\usepackage{subfig}

\title{Linguistic generalization and compositionality in modern
  artificial neural networks}

\author[1,2,3]{Marco Baroni\thanks{Contact information: Department of Translation and Language Sciences (UPF), Carrer Roc Boronat 138, 08018 Barcelona, Spain. Email: \texttt{mbaroni@gmail.com}. ORCID ID: 0000-0001-5066-3580}}

\affil[1]{Catalan Institute for Advanced Studies and Research}
\affil[2]{Department of Translation and Language Sciences, \authorcr
  Universitat Pompeu Fabra}
\affil[3]{Facebook Artificial Intelligence Research}

\begin{document}

\maketitle

\begin{abstract}
  In the last decade, deep artificial neural networks have achieved
  astounding performance in many natural language processing
  tasks. Given the high productivity of language, these models must
  possess effective generalization abilities. It is widely assumed
  that humans handle linguistic productivity by means of algebraic
  compositional rules: Are deep networks similarly compositional?
  After reviewing the main innovations characterizing current deep
  language processing networks, I discuss a set of studies suggesting
  that deep networks are capable of subtle grammar-dependent
  generalizations, but also that they do not rely on systematic
  compositional rules. I argue that the intriguing behaviour of these
  devices (still awaiting a full understanding) should be of interest
  to linguists and cognitive scientists, as it offers a new
  perspective on possible computational strategies to deal with
  linguistic productivity beyond rule-based compositionality, and it
  might lead to new insights into the
  less systematic generalization patterns that also appear in natural language.\\
  \\
  \noindent{}\textbf{Keywords:} artificial neural networks, deep
  learning, linguistic productivity, compositionality
\end{abstract}

\section{Introduction}
\label{sec:intro}

Neural networks have been a prominent tool to model cognitive
phenomena at the mechanistic level since at least the mid eighties
\cite{Rumelhart:etal:1986}. In the last decade, under their ``deep
learning'' re-branding, neural networks have also proven their worth
as astonishingly successful general-purpose, large-scale
machine-learning algorithms \cite{LeCun:etal:2015}. In the domain of
natural language, today neural networks are core components of
effective machine-translation engines such as Google
Translate\footnote{\url{https://translate.google.com/}} and
DeepL.\footnote{\url{https://www.deepl.com/translator}} OpenAI
recently caused controversy when it announced that it would not make
its new language-modeling network publicly available, as it generates
novel text about arbitrary topics that is so realistic and coherent
that it could easily be deployed in malicious applications, such as
bulk creation of faked or abusive
content.\footnote{\url{https://openai.com/blog/better-language-models/}}
The debate on the linguistic abilities of neural networks of the 80s
and 90s involved small experiments and theoretical speculation about
whether neural networks would ever be able to process language at
scale (e.g.,
\cite{Smolensky:1990,Elman:1991,Fodor:Lepore:2002,Pinker:Ullman:2002,Marcus:2003},
among many others). Given the impressive empirical results achieved by
modern neural networks, the interesting question today is not
\emph{whether}, but \emph{how} neural networks achieve their language
skills, and what causes the surprising and sometimes dramatic failures that still
affect them \cite{Lake:etal:2016,Marcus:2018}.

Many early neural networks were developed with the specific
purpose of understanding mental processes, and thus cognitive or
biological plausibility was a central concern. Modern deep networks
are instead optimized for practical goals, such as better translation
quality or information extraction. %
It is thus unlikely that their behaviour will closely mimic human
cognitive processes. I contend, however, that their high natural
language processing performance makes them very worth studying from
the perspective of cognitive science.  Following an early proposal by
McCloskey \cite{McCloskey:1991}, we should treat this as comparative
psychology. Just like the communication systems of primates and other
species can shed light on the unique characteristics of human language
(e.g., \cite{Schlenker:etal:2016,Townsend:etal:2018}), studying how
artificial neural networks accomplish (or fail to accomplish) sophisticated
linguistic tasks can provide important insights on the nature of such
tasks, and the possible ways in which a computational device can (or
cannot) solve them. This is the perspective I adopt here in looking at
linguistic productivity and compositionality in deep networks.

Natural languages are characterized by immense \emph{productivity}, in
the sense that they license a theoretically infinite set of possible
expressions. Linguists almost universally agree that
\emph{compositionality}, the ability to construct larger linguistic
expressions by combining simpler parts, subtends productivity. The
focus is typically on \emph{semantic} compositionality, the principle
whereby the meaning of a linguistic expression is a function of the
meaning of its components and the rules used to combine them
\cite{Frege:1892,Montague:1970a}. When studying the generalization
properties of neural networks, I believe it is more useful to consider
a broader notion of compositionality, also encompassing, for example,
the syntactic derivation rules allowing us to judge the grammaticality
of nonce sentences independently of their meaning
\cite{Chomsky:1957,Chomsky:1965}. Indeed, compositionality is
conjectured to be a landmark not only of language but of human thought
in general
\cite{Fodor:Pylyshyn:1988,Fodor:Lepore:2002,Lake:etal:2016}, and the
compositional abilities of neural networks have been tested on tasks
that are not semantic \cite{Marcus:2003} or even linguistic in nature
\cite{Lake:etal:2015}. If a system is not compositional in this more
general sense, it will not, \emph{a fortiori}, be able to build
complex semantic representations by parallel composition of syntactic
and semantic constituents.

Compositional operations in language (and thought) are argued to
constitute a rule-based algebraic system, of the sort that can be
formally captured by symbolic functions with variable slots. It
follows that compositionality is ``systematic'', in the sense that a
function must apply in the same way to all variables of the right
type. As famously put by Fodor, if you know the correct compositional
rules to understand \emph{``John loves Mary''}, you \emph{must} also
understand \emph{``Mary loves John''} \cite{Fodor:Lepore:2002} (Fodor
and colleagues distinguish systematicity and compositionality:
simplifying somewhat, they see compositionality, in the stricter
semantic sense presented above, as the natural consequence of applying
systematic rules in the domain of natural language). Neural networks
are not thought to be capable of acquiring systematic
rules. Their linguistic generalization abilities have thus been the
focus of much research in the past (see, e.g.,
\cite{Christiansen:Chater:1994,Marcus:1998,Phillips:1998,vanderVelde:etal:2004,Brakel:Frank:2009}
among many others). Note that productivity \emph{per se} does not
entail systematic compositionality. Some forms of generalization
outside language are not rule-based (and not systematic). For example,
similarity-driven reasoning about concept instances is probably too
fuzzy and prototype-based to be accounted for by systematic rules
\cite{Murphy:2002}. One could also imagine a language that is
productive but not (systematically) compositional. For example,
Hockett \cite{Hockett:1960}, reflecting about the origins of language,
conjectured a stage in which new expressions are formed not by
systematic composition of smaller parts, but by blending unanalyzed
wholes in inconsistent ways. Modern languages also exhibit many corners of non-systematic,
partial productivity, a point I'll return to in the conclusion. Still,
systematic composition rules are an extremely powerful generalization
mechanism. Once you know that \emph{super-} attaches to adjectives to
form other adjectives, you can in principle understand an infinite (if
rather contrived) set of words: \emph{super-good},
\emph{super-super-good}, etc. In this context, it has been argued that
lack of compositionality is one reason why modern neural networks, in
striking contrast to humans, require huge amounts of data to induce
correct generalizations \cite{Lake:etal:2016}.


In this article, I would like to introduce researchers interested in
compositionality from a  cognitive perspective to
some relevant recent work about linguistic productivity in
modern deep networks. After briefly reviewing the main novelties
characterizing current deep language-processing architectures, I will
present experimental evidence that these systems are at the same time
able to capture subtle syntactic generalizations about novel forms
(thus handling a sophisticated form of grammatical productivity), and
failing to show convincing signs of rule-based compositionality. I
will conclude with some considerations about the significance of these
results for the general study of linguistic productivity.

\section{Modern deep networks for language processing: what has changed}
\label{sec:modern-deep-networks}

\begin{figure}[p]
  \begin{center}
    \subfloat[\label{sub:original}]{\includegraphics[width=.45\textwidth]{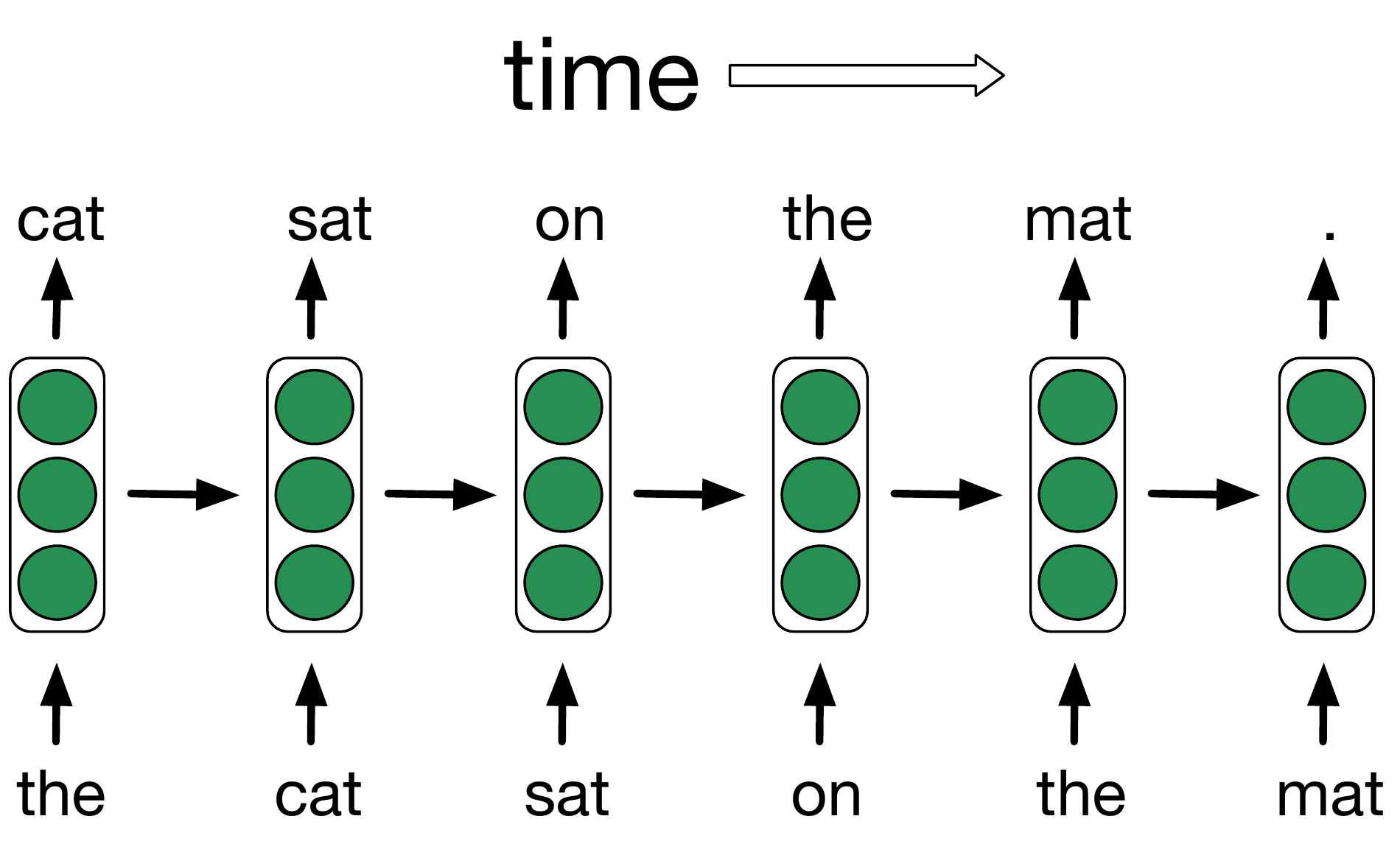}}\hspace{13mm}
    \subfloat[\label{sub:gates}]{\includegraphics[width=3.5cm]{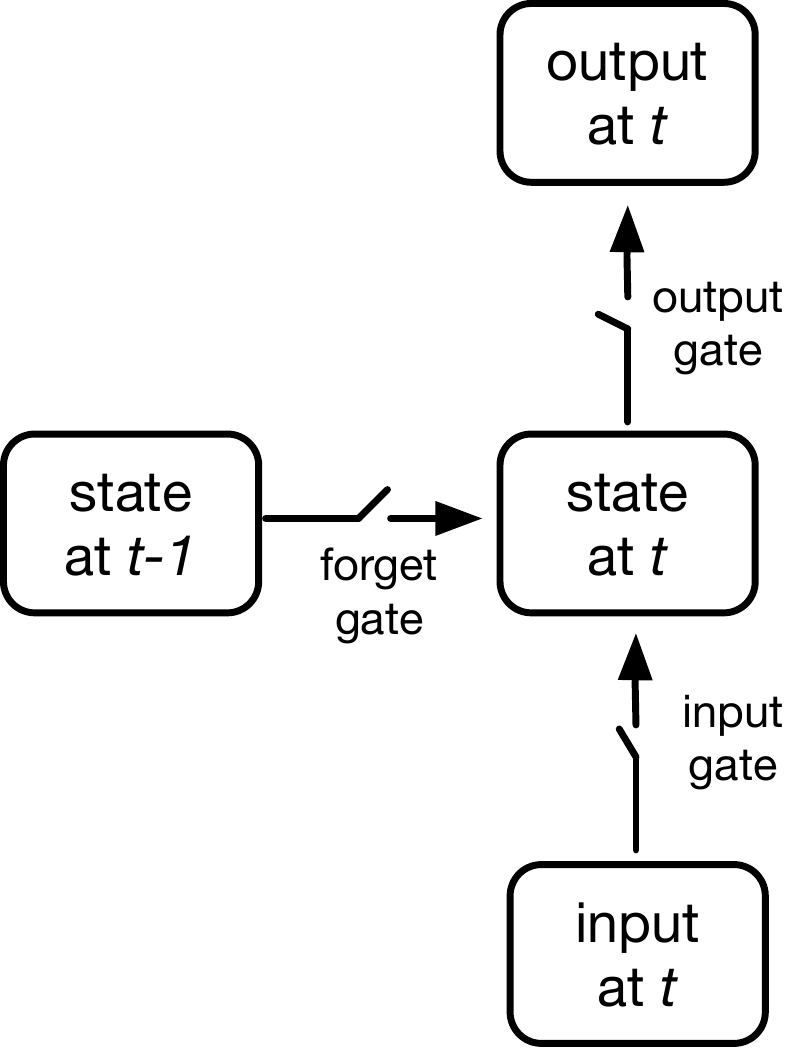}}\\
    \subfloat[\label{sub:encoder_decoder}]{\includegraphics[width=.45\textwidth]{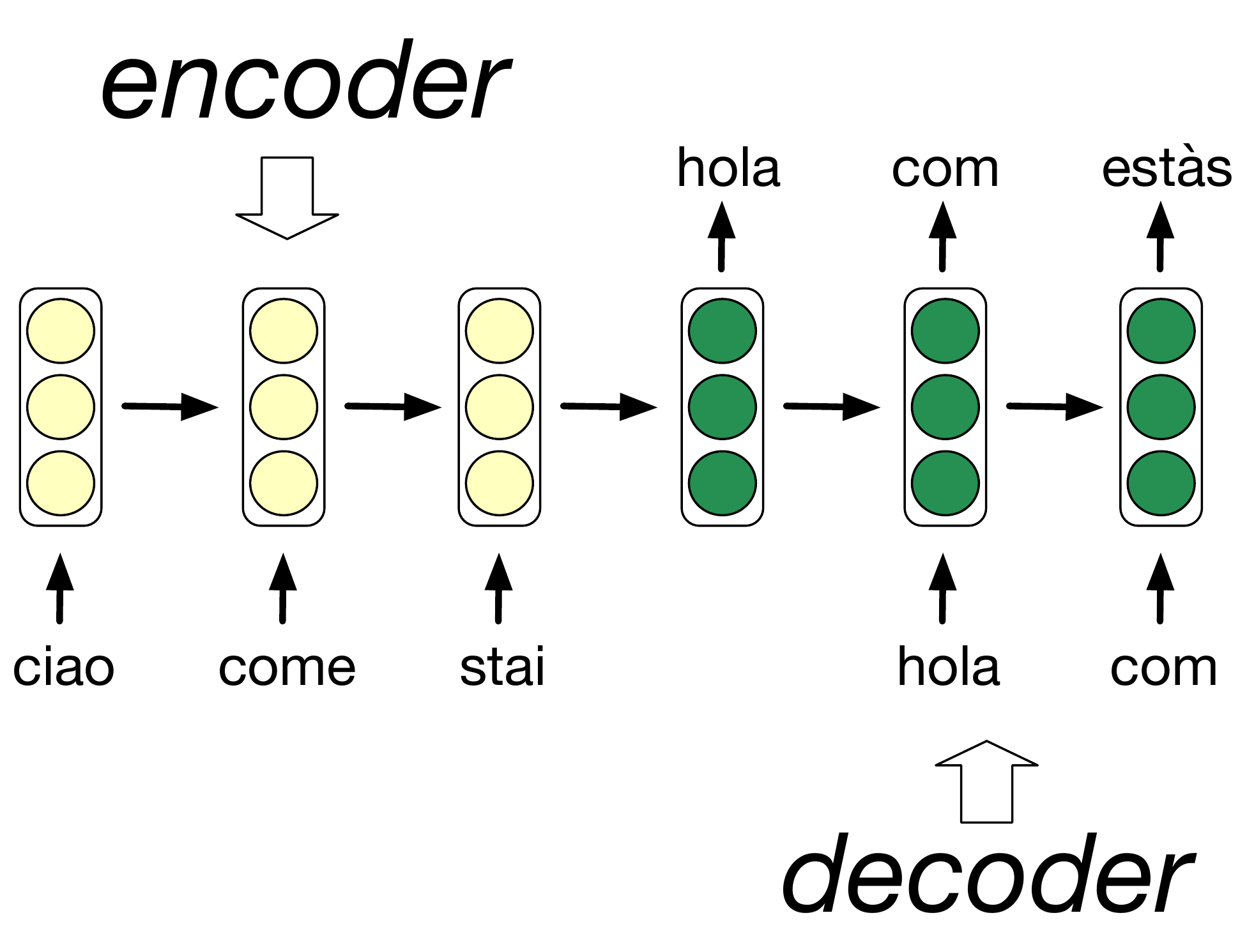}}\hfill
    \subfloat[\label{sub:attention}]{\includegraphics[width=.45\textwidth]{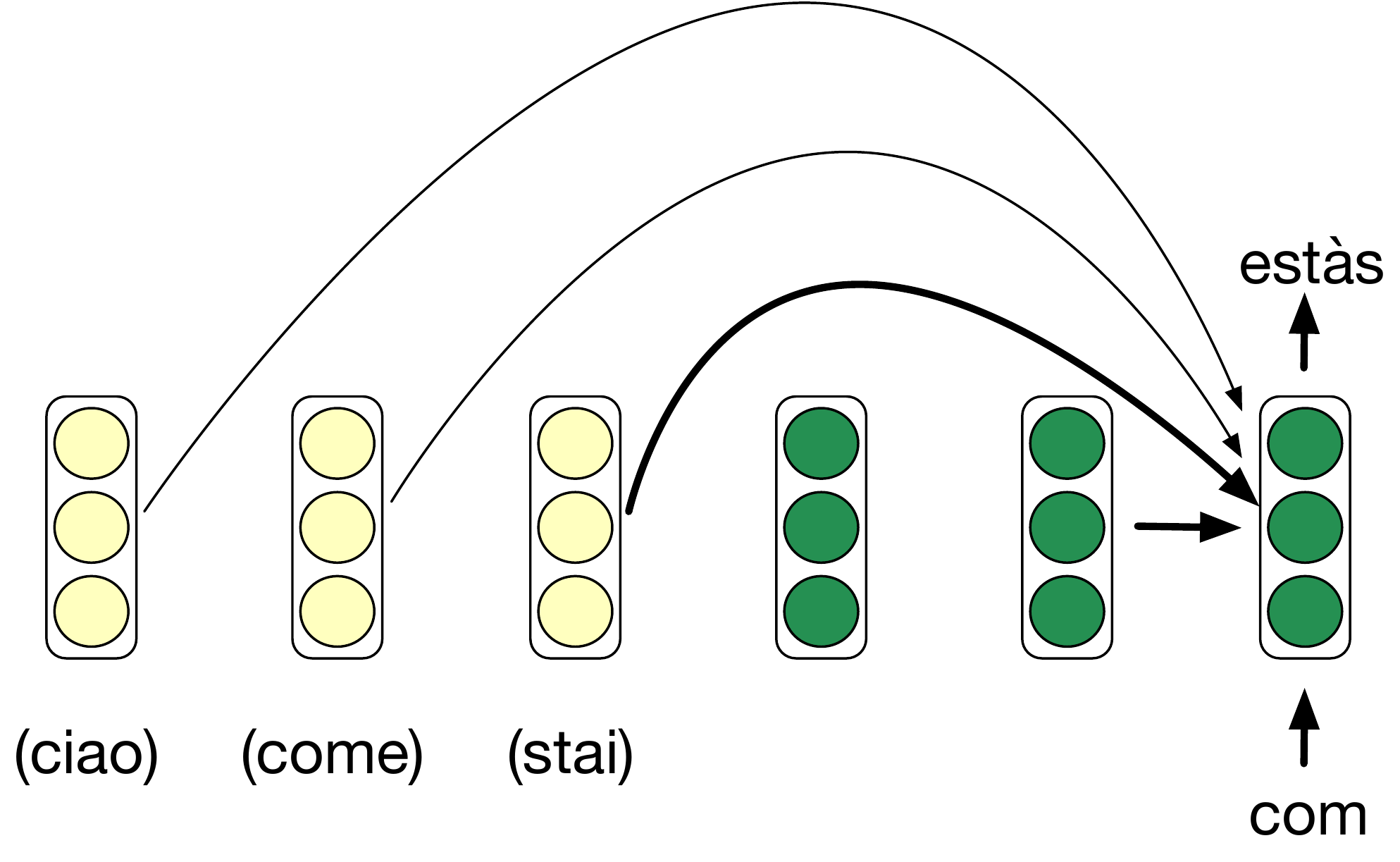}}
    \caption{Architectural features of modern sequence-processing
      networks. (a) A traditional \textbf{recurrent network}
      processing a sequence of words in multiple time steps. The
      arrows represent sets of weighted connections (a single arrow
      stands for multiple unit-to-unit connections). The 3 green
      circles represent the same network at different time steps,
      which might be structured into multiple layers (not depicted in
      the figure). The output at time $t$ is a (non-linear) function
      of the current input as well as the state of the network at time
      $t-1$ (information is carried through time by the
      \emph{recurrent} connections represented by vertical arrows in
      the figure). (b) In modern sequence-processing networks, a set
      of \textbf{gates} modulate the amount of information flowing
      through the connections of the network. (c) The
      \textbf{encoder-decoder} architecture is modular, with separate
      sub-networks (in yellow and green in the figure) trained to
      process input (encoder) and generate output (decoder). The
      decoder is typically initialized from the last state of the
      encoder. (d) In \textbf{attention}-enhanced architectures, the
      state of the encoder at each time step is separately stored in
      memory, and at each decoding step the network dynamically
      determines how much information to read from each of the memory
      slots. The figure schematically depicts the step in which the
      decoder makes a prediction based on the last word it produced
      (\emph{com}), its previous state and attention-mediated
      memorized states from the decoder (with line thickness
      symbolizing the relative weight assigned to each memorized
      state). Unlike the previous diagrams, this one only shows the
      connections that are active in the last depicted processing
      step.}
    \label{fig:rnns}
  \end{center}
\end{figure}


Much of the last-decade improvements in neural network performance are
due to the availability of larger training data sets that, together
with computational power and better optimization methods, enabled
large-scale data-driven training of complex, multi-layer architectures
\cite{LeCun:etal:2015,Goodfellow:etal:2016}. In the domain of
language, large corpora made it possible to train networks with the
simple \emph{language modeling} method
\cite{Graves:2012,Mikolov:2012}. In this setup, the weights of a
sequence-processing network are set by optimizing the objective of
predicting the next word in a text, given the previous context. This
is schematically illustrated for a recurrent network in
Fig.~\ref{fig:rnns}\subref{sub:original}. Nowadays, language modeling
is used as a general-purpose way to pre-train networks to perform
linguistic tasks \cite{Radford:etal:2019}. It is also an interesting
training regime from a cognitive point of view, since humans in many
cultures are also exposed to large amounts of raw language data during
acquisition (e.g., \cite{Landauer:Dumais:1997}), and predicting what
comes next plays a central role in cognition
\cite{Bar:2007,Levy:2008,Pickering:Garrod:2013,Clark:2016}. Of course,
prediction is not the only task humans perform when learning a
language, and how to design more varied and human-like training
environments is an open research issue.

Important advances have also been made in architectural terms. The
original sequence-processing recurrent network schematically
illustrated in Fig.~\ref{fig:rnns}\subref{sub:original} reads some
input (e.g., a word), and produces an output (e.g., a guess about the
next word) at each time step. The output of the network at time $t$ is
a non-linear function of the input at time $t$, as well as of the
state of the network itself at step $t-1$ (weighted by \emph{recurrent
  connections} that propagate activations across time)
\cite{Elman:1990}.

\emph{Gated} recurrent networks, such as long short-term memory
networks \cite{Hochreiter:Schmidhuber:1997} and gated recurrent units
\cite{Chung:etal:2014}, possess mechanisms regulating the dynamics of
information processing across time, whose parameters are jointly
induced with the rest of the network in the training phase. In particular,
the network gates can ``decide'', at each time step and for each unit, how much it
should be updated with information about the current input
(vs.~preserving currently stored information), and how much it should
contribute to (the hidden representation determining) the current
output. Such gating mechanisms, schematically illustrated in
Fig.~\ref{fig:rnns}\subref{sub:gates}, allow longer-term and more
nuanced control of the information flow. Gates have proven empirically
extremely effective, and are standard in modern language-processing
networks \cite{Goldberg:2017}.

Another important innovation consisted in decoupling input and output
processing through \emph{encoder-decoder} architectures
\cite{Sutskever:etal:2014}. As sketched in
Fig.~\ref{fig:rnns}\subref{sub:encoder_decoder}, separate sub-networks
are trained to process the input and generate the output, with the
last state of the first network (the encoder) used to
initialize the second (the decoder). Input-output decoupling
allows effective handling of \emph{sequence-to-sequence} tasks, in
which a sequence (e.g., a sentence in a language) has to be mapped
onto another sequence (e.g., a sentence in another language),
especially where input and output sequences are very different. This
approach is by now standard in machine translation (where, for
example, it permits flexible mapping between languages with different
word orders), but it is extremely general, and it has for example also
been employed to convert linguistic instructions to actions and
sentences to semantic representations
\cite{Dong:Lapata:2016,Mei:etal:2016}.

A (learned) \emph{attention} mechanism, in its original form
\cite{Bahdanau:etal:2015}, automatically allows the decoder to read
more or less information from different encoder states (on
the basis of similarity computations between vectors representing
current and past states). As schematically illustrated in
Fig.~\ref{fig:rnns}\subref{sub:attention}, when it is about to
translate the word following \emph{com} (``\emph{how}'' in Catalan),
an attention-augmented network might decide to read more from the
encoder state corresponding to the word that immediately follows
\emph{come} (``\emph{how}'') in the Italian source.  Attention plays
an increasingly important role in modern encoder-decoder
architectures, to the point that the most successful contemporary
models dispense with recurrent connections altogether, and rely
instead on a rich attention mechanism to keep track of relevant past
information \cite{Vaswani:etal:2017}.

Modern sequence-processing networks are complex systems, equipped with
strong structural priors such as gates, encoding and decoding modules
and attention. They should not be thought of as ``tabulae
rasae'', as they often were in early debates on connectionism. At
the same time, the ``innate'' biases they encode are rather different
from those assumed to shape human linguistic
competence. 
%
Some researchers \emph{are} trying to inject into modern networks
priors closer to those traditionally postulated by linguists, such as a
preference for hierarchical tree structures, (e.g.,
\cite{Socher:etal:2013b,Dyer:etal:2016}; see also Brennan's
contribution to this issue). Models of this latter kind have however not yet proven their
worth as generic language processing devices, and I will not delve
further into them. Intriguingly, \cite{Williams:etal:2018}
recently found that, when such models are not provided with explicit
information about conventional compositional derivations, they come up
with tree structures that do not resemble those posited by linguists
at all. This is in line with the basic tenet of this paper, that neural
networks might solve complex linguistic tasks, but not in
the way we expect them to be solved.




\section{Colorless green grammatical generalization in deep networks}
\label{sec:colorless}

There is no doubt that modern language-processing neural networks can
generalize beyond their training data. Without such ability, their
astounding performance in machine translation
\cite{Vaswani:etal:2017,Edunov:etal:2018} would remain unexplained, as
most sentences, or even long word sequences, in any text to be
translated are extremely unlikely to have ever been produced before
\cite{Baroni:2009}. Recently, there has been widespread interest in
understanding whether this performance depends on shallow heuristics,
or whether the networks are indeed capturing grammar-based
generalizations, of the sort that would be supported by symbolic
compositional rules (e.g.,
\cite{Linzen:etal:2016,Chowdhury:Zamparelli:2018,Wilcox:etal:2018}).

Gulordava and colleagues~\cite{Gulordava:etal:2018} test the networks'
grammatical ``intuitions'' in a setup that is strictly controlled to
insure they are tapping into their productive competence (as opposed
to memorized patterns). They train a gated recurrent network on large
Wikipedia-derived corpora, using the language modeling objective. They
then feed it minimal pairs of sentences respecting/violating
long-distance number agreement. Crucially, the test items are
semi-randomly generated nonsense sentences. For example, one minimal
pair is: ``\emph{I realize the \textbf{wars} on which I should revise
  your hunt \textbf{understand/understands}}'' (here, the plural verb variant with
with \emph{understand} is the grammatical one). The model,
without further task-specific tuning, is tasked with computing the
probability of the two variants, and it is said to have produced the
correct judgment if it assigns a higher probability to the
grammatical one.  Gulordava's nonsensical twist strips off possible
semantic, lexical and collocational confounds, focusing on the
abstract grammatical generalization.

Gulordava and colleagues run the experiment in English, Hebrew,
Italian and Russian. In all cases, neural networks
display a preference for the grammatical sentences that is well
above chance level and competitive baselines (the lowest performance
occurs in English, where the network still guesses correctly 74\%
of the cases, where chance level is at 50\%). Moreover, for Italian, they
compare the network performance to human subjects taking
the same test. Human accuracy turns out to be only marginally above
that of the network (88.4\% vs.~85.5\%).

These results suggest that neural networks capture abstract,
structure-based grammatical generalizations. However, the evidence is
indirect, and others \cite{Kuncoro:etal:2018b,Linzen:Leonard:2018}
have suggested that the networks are really capitalizing on shallow
heuristics (such as: ``percolate the number of the first noun in a
sentence to all verbs''). Lakretz and colleagues
\cite{Lakretz:etal:2019} conducted extensive ablation and connectivity
studies of the Gulordava network. They found that the network
specialized very few units to the task of carrying long-distance
number information. For example, when the activation of these units is
fixed to 0, the network performance on agreement tasks slide towards
chance level. Importantly, these units are strongly connected to a
sub-network of nodes that can be independently shown to be sensitive
to hierarchical syntactic constituency. Unveiling this circuit in the
network suggests that the latter has indeed developed genuine
grammatical processing mechanisms, and it is not simply relying on
surface heuristics when computing agreement.

The kind of productivity that was probed in these studies is
grammatical in nature. Just like in Chomsky's famous ``\emph{colorless
  green ideas}'' example \cite{Chomsky:1957}, Gulordava's network can
tell apart subtly different grammatical and ungrammatical nonsensical
sentences, that are certainly very far from anything it was exposed to
during training. Lakretz' analysis of the network further suggests
that its behaviour relies on genuine sensitivity to grammatical
structure. The traditional linguistic story about a cognitive device
displaying this behaviour would be that it possesses a compositional
rule-based system, allowing it to reliably process novel linguistic
input (such as Gulordava's stimuli). We have no evidence, yet, about whether Gulordava's network
possesses something akin to such system, but we will now turn to
experiments with similar networks directly probing their
compositionality through a miniature language designed for this purpose,
that suggest that they don't.

\section{Compositional generalization: can deep networks dax twice?}

Lake and Baroni \cite{Lake:Baroni:2017} introduced SCAN, a benchmark
to test the compositional abilities of sequence processing networks,
later extended by \cite{Loula:etal:2018}. The SCAN miniature language
is characterized by a grammar generating a large but finite number of
linguistic navigation commands, and an interpretation function
associating a semantic representation (a sequence of action symbols)
to each possible command. The primitives of the language are verbs
such as \emph{jump} and \emph{run}, mapped to the corresponding
actions (e.g., JUMP, RUN). Primitives are combined with a set
of adverb-like modifiers and conjunctions, resulting in composite
expressions denoting action sequences. For example, if [[x]] is
the action associated to expression ``x'' by the interpretation
function, then ``x \emph{and} y'' maps to [[x]] [[y]] and
``x \emph{twice}'' maps to [[x]] [[x]]. Consequently,
\emph{``jump twice and run''} is compositionally mapped to the action
sequence: JUMP JUMP RUN.

The general evaluation paradigm is as follows. An encoder-decoder
network is trained on a set of SCAN commands for long enough that the
network learns to accurately execute them (that is, to map them to the
corresponding action sequences). The network performance is then
evaluated on executing a set of test commands that were not encountered during
training. Note that, unlike in the experiments reviewed in the
previous section, where it had to assign a probability to
pre-determined sentences, here the network has to actively produce an
output action sequence, thus its generative abilities are more directly
probed.

By splitting the possible SCAN commands into different training and
testing partitions, we can gain insights into what the network is
actually learning. I will focus here on the results obtained with 3 of
the proposed splits, and their implications.  In the \emph{random}
split, 80\% of the commands are used for training, the remaining 20\%
for testing. This split checks the network ability to handle generic
productivity (of the sort that might occur in standard
machine-translation benchmarks), since all test expressions are new
for the network. However, there is no \emph{controlled} difference
between the two sets, and the network will in general have seen a
number of examples quite similar to those it has to execute. For
example, the test set contains the command \emph{``look around left
  twice and jump right twice''}. This is new, but the training data
contain examples of ``\emph{look around left twice}'' and ``\emph{jump
  right twice}'', both on their own and in conjoined expressions
(e.g., \emph{``\textbf{look around left twice} and turn left''},
\emph{``run twice and \textbf{jump right twice}''}).

In the \emph{jump} split, the training set contains all possible
commands with all primitives except \emph{jump}. During training,
\emph{jump} is presented multiple times, but only in isolation. The
test set is then made of all composite commands containing
\emph{jump}. For example, at training the network is exposed to
\emph{``run twice''} and \emph{``walk and look''}, and at test time it must
execute \emph{``jump twice''} and \emph{``walk and jump''}. The split is
straightforward for a system possessing composition rules such as: ``x
\emph{twice}'' maps to [[x]] [[x]]. This is akin to a human subject
learning a new verb \emph{daxing}, and being immediately able to
understand what \emph{``dax twice''} means.

However, since \emph{jump} only occurs in isolation at training time,
the tested specimen could also reasonably conclude that the latter has
a different distribution from the other verbs, and refuse to
generalize it to novel composite contexts
\cite{Wonnacott:etal:2008}. Loula and colleagues
\cite{Loula:etal:2018} introduce different partitions that control for
this factor. In particular, in the \emph{around-right} split, the
training data contain all possible commands, except those where
\emph{around} is combined with \emph{right}. Still, the network is
given plenty of distributional evidence that \emph{right} and
\emph{left} function identically otherwise, and it is exposed to many
examples illustrating the behaviour of \emph{around} in combination
with \emph{left}. The training data contain, a.o., the commands
\emph{``run left''}, \emph{``run right''}, \emph{``jump opposite
  left''}, \emph{``jump opposite right''}, \emph{``look around
  left''}. The test data include \emph{``look around right''}.

The results in the original papers, and the further experiments of
\cite{Bastings:etal:2018} with more carefully-tuned models, tell a
simple story. Modern recurrent sequence-processing networks, just like
conjectured by Jerry Fodor and Gary Marcus, are able to generalize in
a fuzzy, similarity-based way that allows them to succeed in the
\emph{random} split (100\% average test accuracy and s.d.~$\approx$
0.0\% across multiple runs with different initializations). However,
they utterly fail at the \emph{jump} and \emph{around-right} splits,
that require inducing systematic compositional rules (12.5\% accuracy
with 6.6\% s.d.~and 2.5\% accuracy with 2.7\% s.d., respectively).

In very recent work, Dess\`{i} and Baroni \cite{Dessi:Baroni:2019}
found a somewhat more intriguing pattern. They replaced the gated
recurrent network architectures used in earlier SCAN work with an
out-of-the-box convolutional network that dispenses with recurrence by
heavily relying on attention, and that has independently been shown to
achieve competitive results in machine translation
(\cite{Gehring:etal:2017}; Dess\`{i} and Baroni were not able, for
the time being, to trace the difference in performance between this
network and the previously tried models back to their architectural
differences). The new model is still able to perfectly generalize in
the random split (100\% test accuracy, s.d.~$\approx$ 0.0\%), but now
it reaches a surprising middle ground in the ``compositional'' splits:
69.2\% accuracy (8.2\% s.d.)~with \emph{jump} and 56.7\% (10.2\%
s.d.)~with \emph{around-right}. As chance level accuracy in these
tasks is practically 0\%, the results show that the network does get
some important generalizations right. Still, if it extracted the
correct rules, we would expect it to be perfectly accurate, which is
not the case. Dess\`{i} and Baroni initially conjectured that the
network learned a subset of compositional rules, explaining its
partial success. For example, it could be that the network learned how
to map ``x \emph{twice}'' to [[x]] [[x]], but failed to learn the
corresponding ``x \emph{thrice}'' rule. However, a follow-up error
analysis showed this not to be the case. The network makes errors
relatively uniformly across composition frames, and, qualitatively, it
does not display any trace of systematicity. For example, in the
\emph{jump} split the network executes \emph{``jump left after walk''}
correctly, but fails \emph{``jump left after run''}. In the
\emph{around-right} split, the network can execute \emph{``run around
  right''}, but not \emph{``walk around right''}.

Similar conclusions were drawn by Andreas \cite{Andreas:2019} from a
very different experiment. Andreas trained sender and receiver
sequence-processing networks to play a ``communication game'': The
sender must describe the properties of a set of objects to the
receiver through a discrete communication channel, and the receiver
must reconstruct the correct target objects. The agents were trained
by rewarding successful communication. Across random initializations,
the sender agent developed more or less compositional codes for its
messages to the receiver. Interestingly, at least some codes with low
degrees of compositionality were as good at generalizing to new object
descriptions as more compositional ones. Again, neural networks can be
productive without being compositional.

\section{Conclusion}

When critics of classic connectionism argued that neural networks are
intrinsically incapable to induce symbolic composition rules, they
probably assumed that this would severely limit their practical
ability to handle natural language. The empirical evidence concerning
modern deep networks is surprising, as it suggests that they are
extremely proficient at language, while indeed not being
compositional. Note in particular that the experiments by Gulordava,
Lakretz and others I reviewed in Section \ref{sec:colorless} above
suggest that the linguistic proficiency of neural networks extends
beyond shallow pattern recognition, to competence about
structure-dependent generalizations of the sort traditionally
attributed to the command of systematic compositional rules. Our
current understanding of the strategies learned by these networks is
very limited, and our highest priority should be to develop better
analytical tools to uncover the mechanisms that lead to the detected
dissociation of productive grammatical competence  and
systematic compositionality.

From the perspective of AI research, one central question is whether
making neural networks more compositional, for example by means of
more structured modular architectures \cite{Andreas:etal:2016b}, will
also make them more adaptive and faster at learning, while not costing
in terms of generality. Current deep networks are also brittle in
surprising ways. For example they are easily fooled by ``adversarial''
examples (e.g., words with a few characters shifted) that would be
trivially handled by humans
\cite[e.g.,][]{Belinkov:Bisk:2018,Ebrahimi:etal:2018,Zhao:etal:2018}. Explicitly
compositional architectures might provide added robustness to similar
attacks, or at least afford better insights into the often mysterious
failings of the networks. In turn, this might lead to progress in
ambitious natural language processing tasks where the success of
modern deep networks are less clear-cut, such as machine reading and
natural language inference
\cite[e.g.,][]{Jia:Liang:2017,McCoy:etal:2019}.

Still, the way in which current models generalize without possessing
anything resembling compositional rules also offer an intriguing
opportunity for comparative studies to linguists and cognitive
scientists. Classic and modern criticism of neural networks emphasizes
the aspects of human language that are best characterized by
clear-cut, algebraic rules. Language, however, is also host to plenty
of productive phenomena that obey less systematic, fuzzier laws,
ranging from phonologically-driven generalizations of irregular
inflections \cite{Albright:Hayes:2003}, to partial semantic
transparency in derivational morphology \cite{Marelli:Baroni:2015}, to
semi-lexicalized constraints in syntax
\cite{Goldberg:Jackendoff:2004}, to the early stages of
grammaticalization in language change
\cite{Hopper:Traugott:2003}. Progress in understanding the linguistic
capabilities of neural networks might help us to make precise
predictions about the origin, scope and mechanics of these phenomena,
and ultimately to develop a more encompassing account of the amazing
productivity and malleability of human language.


\section*{Acknowledgments}

I would like to thank Gemma Boleda, Roberto Dess\`{i}, Diane
Bouchacourt, Emmanuel Dupoux, Kristina Gulordava, Dieuwke Hupkes,
Douwe Kiela, Jean-R\'{e}mi King, Germ\'{a}n Kruszewski, Tal Linzen,
Tomas Mikolov, Paul Smolensky, Matthijs Westera, the NTNU
\emph{``Towards mechanistic models of meaning composition''} workshop
participants, my colleagues at FAIR, my co-authors, the reviewers for
the Philosophical Transactions of the Royal Society and especially,
Brenden Lake and Yair Lakretz for many enlightening discussions about
compositionality and neural networks.

\section*{Competing interests}

I have no competing interests.


\end{document}